\documentclass[pdflatex, iicol]{sn-jnl}
\usepackage{subfig}

\usepackage{graphicx}%
\usepackage{multirow}%
\usepackage{amsmath,amssymb,amsfonts}%
\usepackage{amsthm}%
\usepackage{mathrsfs}%
\usepackage[title]{appendix}%
\usepackage{xcolor}%
\usepackage{textcomp}%
\usepackage{manyfoot}%
\usepackage{booktabs}%
\usepackage{algorithm}%
\usepackage{algorithmicx}%
\usepackage{algpseudocode}%
\usepackage{listings}%

\author*[1]{\fnm{Omer} \sur{Subasi}}\email{omer.subasi@pnnl.gov}

\author[1]{\fnm{Oceane} \sur{Bel}}\email{obel@pnnl.gov}

\author[1]{\fnm{Joseph} \sur{Manzano}}\email{joseph.manzano@pnnl.gov}

\author[1]{\fnm{Kevin} \sur{Barker}}\email{kevin.barker@pnnl.gov}

\affil*[1]{\orgdiv{High Performance Computing Group}, \orgname{Pacific Northwest National Laboratory}, \orgaddress{\street{902 Battelle Blvd}, \city{Richland}, \postcode{99354}, \state{WA}, \country{USA}}}

\begin{document}

\title[The Landscape of Modern Machine Learning]{The Landscape of Modern Machine Learning: A Review of Machine, Distributed and Federated Learning}

\abstract{
With the advance of the powerful heterogeneous, parallel and distributed 
computing systems and ever increasing immense amount of data, machine learning has become an indispensable part of cutting-edge technology, scientific research and consumer products. In this study, we present a review of modern machine and deep learning. We provide a high-level overview for the latest advanced machine learning algorithms, applications, and frameworks. Our discussion encompasses parallel distributed learning, deep learning as well as federated learning. As a result, our work serves as an introductory text to the vast field of modern machine learning.
}


\keywords{Machine Learning, Distributed Machine Learning, Deep Learning, Federated Learning, Parallel and Distributed Computing.}

\maketitle

\section{Introduction}
Over the last decade, 
Machine Learning (ML) has been applied to ever increasing immense amount of data that is becoming available as 
more people become daily users of internet, mobile and wireless networks. 
Coupled with the significant advances in deep learning (DL), ML has found more complex applications: from medical to machine translation and speech recognition, to intelligent object recognition, and to smart cities \cite{dong2021survey, sarker2021deep}. 
Modern parallel and heterogeneous computing systems \cite{9286149, 8916327, 9739030}  have enabled such applications by supporting highly parallel training.
These large-scale and distributed systems therefore have become the backbone of modern ML \cite{SurveyonLargeScale, SurveyonDistributed, Large-scale-machine-industrial-settings}.

Federated Learning (FL), as a sub-field of DL, has emerged as a distributed learning solution to provide data privacy \cite{DBLP:conf/aistats/McMahanMRHA17}. Ever since its inception \cite{DBLP:conf/aistats/McMahanMRHA17}, FL has been studied extensively and adapted widely \cite{FromDistributedtoFederated, zhang2021survey, 9599369, 9415623}.

In this study, we review the current landscape of modern ML systems and applications, and offer an overview as a self-contained text. While there are many surveys on large-scale \cite{SurveyonLargeScale, Large-scale-machine-industrial-settings}, distributed ML \cite{SurveyonDistributed}, DL \cite{dong2021survey, sarker2021deep, dargan2020survey}, and FL \cite{FromDistributedtoFederated, zhang2021survey, 9599369, 9415623}, we instead provide a high-level joint view of modern parallel and distributed ML and FL. In this way, our work differentiates itself from the existing literature. In brief, our study 
\begin{itemize}
    \item presents the concepts and methods of ML and DL.
    \item discusses the parallelism and scaling approaches of large-scale distributed ML. Moreover, it explores the communication aspects, such as costs, topologies, and networking, of parallel and distributed training and inference.
    \item introduces FL, its applications and aggregation methods. It then elaborates on the security and privacy aspects as well as the existing platforms and datasets.
    \item summarizes open research questions in the modern landscape of parallel and distributed ML, DL and FL.
\end{itemize}

\begin{figure*}
        \centering
        \includegraphics[width=\linewidth]{
        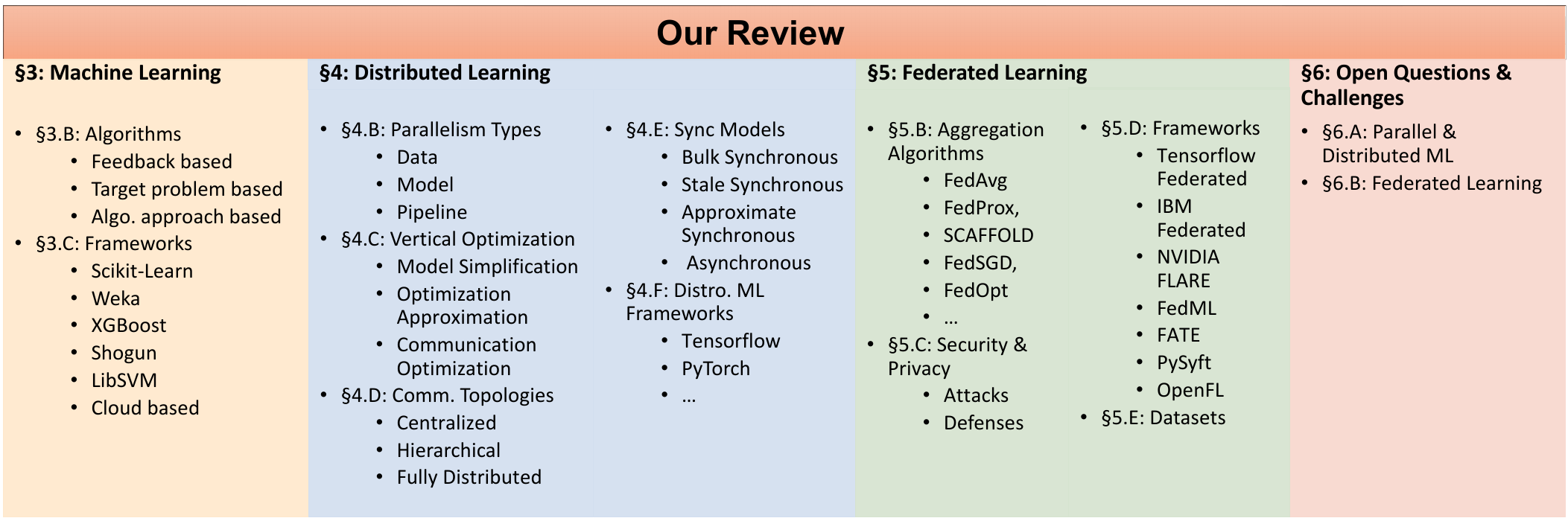
        }
    \caption{The outline of our review.}
    \label{fig:outline}
\end{figure*}

Figure \ref{fig:outline} outlines and summarizes our study.

Our study is organized as follows: 
Section \ref{related} overviews the related work on large-scale and distributed ML.
Section \ref{ml} provides the background on ML.
Section \ref{distributedml} discusses distributed ML.
Section \ref{federatedml} presents FL.
Section \ref{questionschallenges} summarizes the existing open challenges.
Finally, Section \ref{conclusion} concludes our review.

\section{Related Work}
\label{related}
Surveys pertaining to parallel, distributed and large scale ML have been very numerous in the literature \cite{SurveyonLargeScale, SurveyonDistributed, Large-scale-machine-industrial-settings}. Our work is different and unique because it provides an introductory review of the latest joint landscape of ML, DL and FL.

Different than the general surveys such as \cite{sarker2021machine, SurveyonLargeScale, SurveyonDistributed, Large-scale-machine-industrial-settings}, some surveys offer in depth cost and comparisons of algorithms and methods both theoretically and empirically \cite{DemystifyingParallelandDistributed, DistributedComparative}.

Many studies focus on distributed DL. Some of them are \cite{dong2021survey, sarker2021deep, DistributedDeepLearning, ScalableDeepLearningDistributed, dargan2020survey, 9352033FederatedSurveyIEEE}. Moreover, there exists a significant number of surveys that focus on specific types of models such as \cite{shao2022distributed} for graph neural networks (GNNs), \cite{9534784IoT} for Internet-of-Things (IoTs), \cite{9446488Wireless} for wireless networks, \cite{Survey5G} for mobile and 5G networks or for specific target environments such as \cite{ding2023distributed} for unmanned aerial vehicles (UAV).

FL literature unsurprisingly offers many surveys.
Some of the latest surveys are \cite{FromDistributedtoFederated, zhang2021survey, 9599369}.
Among studies having specific topics,
\cite{MOTHUKURI2021619} surveys privacy and security methods for FL, \cite{qu2022blockchain} discusses block chain-based FL.
\cite{el2022differential} presents differential privacy for FL. 
\cite{9415623} offers a survey of FL for IoT.

\section{Machine Learning (ML)}
\label{ml}
In this section, we first overview ML in terms of concepts and goals.
Then we review various ML algorithms.
Finally, we discuss the existing modern ML frameworks.

\subsection{Introduction to ML}
ML is the process of learning from data to perform complex tasks
for which there is no known deterministic and algorithmic solution, or 
building such a solution is not practical. For instance, developing a deterministic 
algorithm based on rules to detect
spam emails is highly impractical. It is not possible to know
the exact list of the detection rules. In addition, these rules most often change over time.
Since the list of the rules may be ever-increasing and even contradictory, 
the maintenance of such algorithms would require constant labor.

\begin{figure}[ht]
    \centering
    \includegraphics[width=\linewidth]{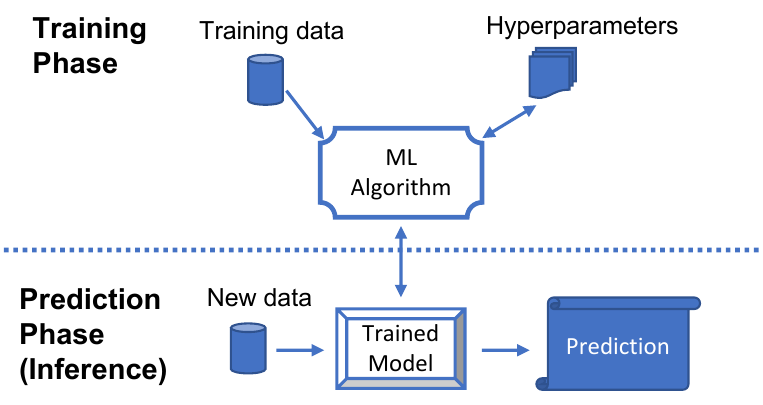}
    \caption{ML phases: Training and prediction (inference).}
    \label{fig:mltraininginfer}
\end{figure}

The ML process is mainly two-fold: Training and prediction (inference).
In the training phase, the parameters of a learning model are optimized based on data.
In the prediction phase, the trained model is deployed to perform predictions on new data.
While in most cases the training and prediction phases are mutually exclusive, in
incremental learning cases, they are coupled together. 
The models in these cases are continuously trained and make predictions.
Figure \ref{fig:mltraininginfer} visualizes the training and prediction phases.

\begin{figure}[ht]
    \centering
    \includegraphics[width=\linewidth]{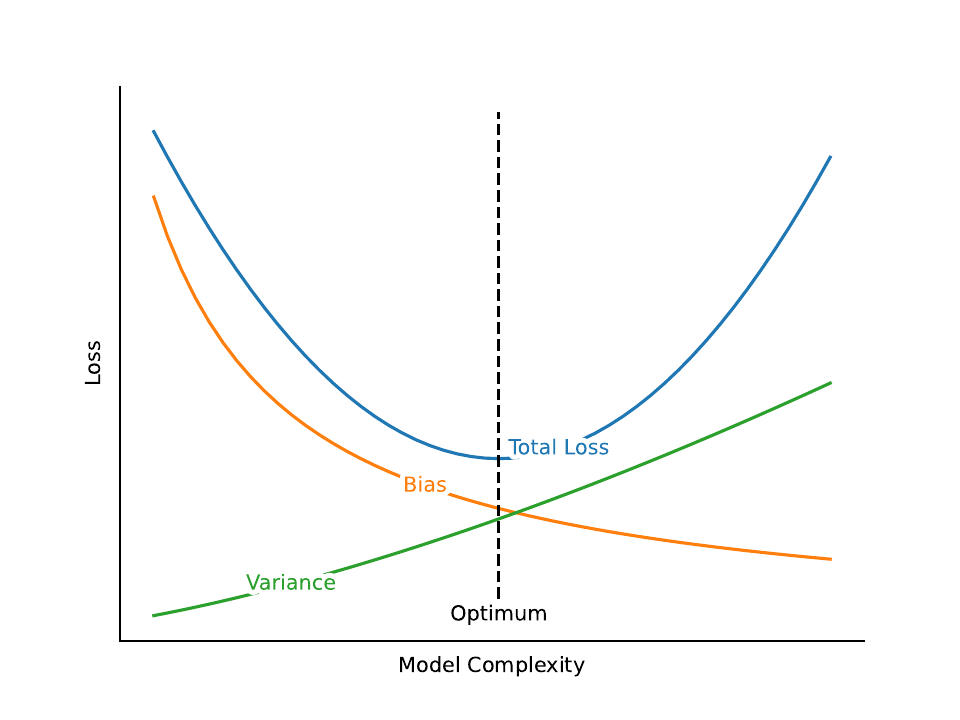}
    \caption{Bias-variance trade-off. Model complexity with respect to bias and variance.}
    \label{fig:biasvariance}
\end{figure}

The main goal of ML is to generalize such that it performs well with unseen data. However, this goal contradicts its optimization goal in which ML tries to minimize the training loss with the training data. As a result, the well-known bias-variance problem emerges. If an ML model over-fits the training data, that is, having high variance, it performs poorly with the unseen data.
On the other hand, if the model under-fits, that is, having high bias, it does not learn important patterns or regularities in the data. Over-fitting typically happens when a model is too complex for the underlying problem. In contrast, under-fitting happens when the model is too simple. Figure \ref{fig:biasvariance} depicts the bias-variance trade-off.


In the following, we present different types of ML tasks. After that, we look into different problems that ML can solve.
Then, we review widely used ML algorithms and methods. 
Finally, we survey the existing ML platforms that are not supported with specialized hardware and not suited for DL or FL.

\begin{figure*}[!htb] 
  \centering
  \includegraphics[width=0.7\linewidth]{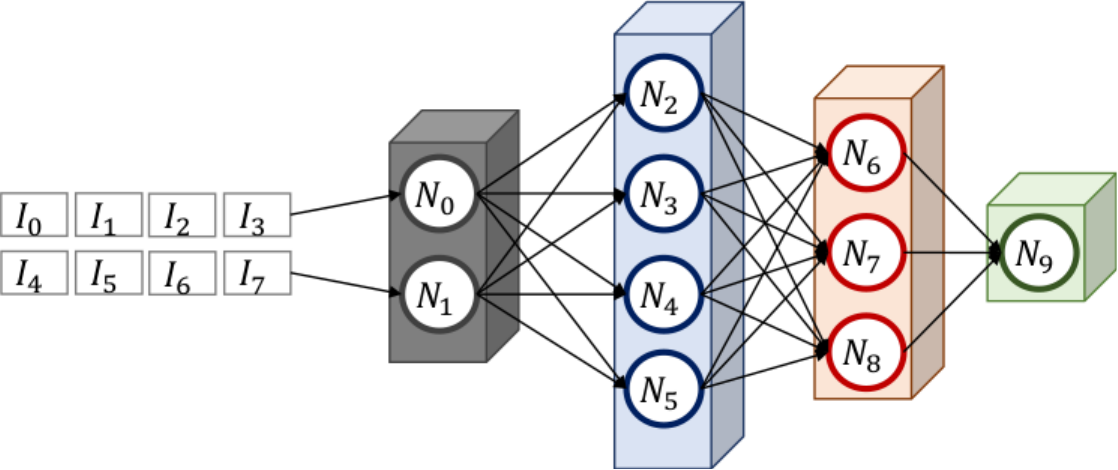}
  \caption{An artificial neural network example.}
  \label{fig:dnnex}
\end{figure*}

\subsection{ML Algorithms}
 ML algorithms can be categorized by the format and requirements of data (external feedback), 
by the type of problems they are designed for (target problem), and by the techniques they use (algorithmic approaches).

It is worth noting that there is another way of categorizing ML: online and offline. In offline learning, the entire training data is available prior to training. This is the most common application of ML. In online learning \cite{hoi2021online}, either the entire data is not available beforehand or it is computationally infeasible to perform training over the entire data at once. An example of the former is sequential training such as time series analysis in financial markets. An example of the latter is learning with a very large dataset which does not fit into the memory and consequently, training becomes prohibitive.

\subsubsection{External feedback}
ML algorithms can be classified based on the external feedback as follows:

Supervised Learning:
Learning is performed by feeding labeled input data 
so that a model's parameters are optimized. 
Labeled data can be desired classes, categories, or numerical outputs corresponding to the training instances.
During training, the optimization is achieved by minimizing a predetermined cost function.
After training, the trained model is deployed to predict
the outputs of new instances. 
An example supervised learning is to classify newly seen handwritten digits by training with the labeled digits.

Unsupervised Learning:
The goal of unsupervised learning is to find structures and patterns in unlabeled data. This means that in  unsupervised learning, the data does not possess desired outputs.
As an example of unsupervised learning, clustering aims to find similar groups (clusters) in given data.
Dimensionality reduction is another example of unsupervised learning where the goal is to find a subset of key features that describes the data well. 

Semi-supervised Learning:
In semi-supervised learning, the amount of labeled data is small while the amount of
unlabeled data is large. Clustering algorithms are typically used to propagate existing
labels to the unlabeled data. An assumption of semi-supervised learning is that similar data shares the same label.

Reinforcement Learning:
Reinforcement learning is applied when an agent interacts with an environment. 
Based on the observations it makes, the agent takes actions. The actions are rewarded or 
penalized according to a reward function. Applications of reinforcement learning lie in the fields such as game theory, robotics and industrial automation.

\subsubsection{Target Problem}
Under this categorization, ML algorithms are grouped according to the kind of problems they are designed to solve.

In classification problems, the aim is to correctly categorize data instances into the known classes.

In regression problems, the goal is to estimate the value of a variable based on other variables (features).

Clustering finds the distinct groups of similar data instances based on a selected similarity metric.

Anomaly and novelty detection is used to find
data instances that are significantly different than others. These instances are called outliers. In anomaly detection, training data consists of both outliers and regular (expected) data instances.
In novelty detection, on the other hand, the goal is used to detect unseen data where training data is free of outliers.

Dimensionality reduction is used to reduce the number features of the training data. In dimensionality reduction,  if a subset of the original set of the features is selected, it is called feature selection. In contrast, 
if features are combined into new ones, it is called feature extraction.
Dimensionality reduction can also be used to decrease computational costs of training. Furthermore, it can also be used to prevent over-fitting. The problem of over-fitting with high-dimensional data is famously known as the curse of dimensionality. The curse of dimensionality arises due to data sparsity in high dimensional spaces.

\subsubsection{Algorithmic Approaches}
ML algorithms can be categorized based on algorithmic approaches that they employ.

Stochastic Gradient Decent (SGD) based algorithms are optimized based on a loss function of the outputs of
the model parameters in the opposite direction of the gradient.
Because at each training step a random subset of data is used, this optimization method is 
called stochastic. 
Many common ML algorithms are optimized with SGD such as artificial neural networks.

Support Vector Machines (SVMs) \cite{hearst1998support} are typically used when the input data is not linearly separable in its original space. They 
map the input data to high dimensional spaces where it becomes linearly separable.
SVMs can be used for classification, regression, and novelty detection.

Artificial Neural Networks (ANNs) are constructed by multiple layers of nodes (neurons) that have inputs, outputs, corresponding feature weights, and an activation function. Layers can be input, hidden, and output layers. ANNs have recently been very successful in tasks such as image classification, object detection, and natural language processing. Figure \ref{fig:dnnex} depicts an example of an ANN.
Some well-known types of ANNs include: 
\begin{itemize}
    \item Convolutional Neural Networks (CNNs) \cite{khan2020survey} are deep neural networks that incorporate convolutions and pooling. While convolutions help with learning local data, pooling help with learning abstract features. CNNs have been extremely successful in tasks such as image classification, object detection, and image segmentation.  
    \item Recurrent Neural Networks (RNNs) \cite{yu2019review} maintains a temporal state of sequence data.
    The temporal state may hold short-term or long-term memory. RNNs are used in tasks such as time series forecasting, natural language processing, and anomaly detection.
    \item Autoencoders \cite{bank2020autoencoders} are ANNs that learn latent representations of input data with no supervision. They are used for dimensionality reduction and visualization of high dimensional data.
    \item Generative Adversarial Networks (GANs) \cite{goodfellow2020generative} are (originally unsupervised) neural networks used to generate data based on a game between a generator and discriminator network. They have been successfully applied in supervised and semi-supervised learning.
    \item Graph Neural Networks (GNNs) \cite{wu2020comprehensive} are a type of ANNs designed to perform learning and prediction on data described by graphs. GNNs provide an easy way to do node, edge, and graph level ML tasks.
    \item Self-Organizing Maps (SOMs) \cite{van2012self} are neural networks which produce a low dimensional
    representation of high dimensional data. SOMs are used for visualization, clustering, and classification. The training is unsupervised where after random initialization,
    neurons compete against each other. 
    \item Boltzmann Machines \cite{hinton1984boltzmann, hinton1986learning} are fully connected ANNs which, unlike other ANNs, have probabilistic activation functions. Neurons output 1 or 0 based on
    Boltzmann distribution. Boltzmann Machines can be used for classifying, denoising, or completing images. 
    \item Deep Belief Networks \cite{hinton2009deep} are stacked Boltzmann Machines designed to tackle larger and more complex learning challenges. They are used for semi-supervised learning.
    \item Hopfield Networks \cite{hopfield2007hopfield, ramsauer2020hopfield} are fully connected networks that are used for tasks such as character recognition.
\end{itemize}

Transformers \cite{lin2022survey} are a class of DL models that has shown extraordinary success in many ML fields including natural language processing and computer vision. Transformers were first introduced by a landmark paper from Google \cite{vaswani2017attention} which were based on a novel mechanism called \textit{Attention}. At its core, a transformer is an encoder-decoder model. The success of Transformers has become a regular news-headliner such as the release of GPT-4 \cite{openai2023gpt4} and ChatGPT \cite{chatgpt}. 

Rule-based algorithms \cite{weiss1995rule} use a set of rules to learn patterns from the input data. They are typically easier to interpret than other ML algorithms.
Decision trees are the most well-known rule-based algorithms.

Evolutionary algorithms \cite{Evolutionary}
 use ideas from biological evolution.
In evolutionary algorithms, the target problem is represented by a set of properties.
The performance metric is called fitness function.
Based on fitness scores, the set of properties is mutated and crossed over.
These algorithms iterate until accurate estimates are obtained.
Evolutionary algorithms can also be used to create other algorithms such as neural networks.

Semantic and Topic algorithms \cite{alghamdi2015survey, 10.4108/eai.13-7-2018.159623} are used to learn specific semantic patterns and distinct relationships in the input data.
An example application of these algorithms is to find the topics and relate them to each other in a given set of documents.

Ensemble algorithms combine other algorithms to obtain a  solution that performs better than the individual algorithms. Different ways to build ensembles are:
\begin{itemize}
    \item Bagging combines multiple classifiers and uses voting to determine the final output.
    \item Boosting is a technique that trains the subsequent models with the data instances misclassified by the preceding models in the chain.
    \item Stacking is the process where a model trains with the outputs of the preceding models in a chain of several models. Stacking typically reduces the classification variance.
    \item Random Forests combine multiple decision trees and output an (weighted) average of the outputs of the individual trees. 
\end{itemize}

\subsection{Existing ML Frameworks}
In this section, we present the existing ML platforms that are not supported with specialized hardware and typically not suited for DL or FL. We then briefly mention the popular ML services in the cloud.

Scikit-Learn \cite{Scikit} is the most popular open-source Python library that offers an extensive suite of ML algorithms. The library is very well maintained and provides a comprehensive set of algorithms, methods, pre-processing, pipelining, model selection and hyper-parameter search capabilities.
It provides interfaces to work with NumPy and SciPy packages.

Weka \cite{weka} is a general-purpose and popular Java ML library. It provides a large collection of algorithms and visualization tools. Weka supports numerous tasks such as pre-processing, classification, regression, clustering and visualization.

XGBoost \cite{chen2016xgboost} is a scalable and distributed gradient boosting library based on decision trees. It implements parallel ML algorithms for classification, regression and ranking tasks.

Shogun \cite{SHOGUN} is a research-oriented open-source ML library. It offers a large number of ML algorithms and cross-platform support by providing bindings with other languages and environments such as Python, Octave, R, Java. Shogun's core library is implemented in C++.

LibSVM \cite{chang2011libsvm} is a specialized C/C++ library for SVMs. It provides interfaces for Python, R, MATLAB and many others.

Many companies offer standard ML and distributed ML services. Moreover, these services often include the support for GPUs and other ML specific hardware. Popular cloud ML services are
Google's Cloud \cite{googleml}, Microsoft Azure \cite{microsoftml}, Amazon's SageMaker \cite{amazonml} and the IBM Watson Cloud \cite{ibmml}.

\section{Distributed Machine Learning}
\label{distributedml}
In this section, we introduce large-scale distributed ML.
We then explore different types of parallelisms used in distributed training. 
Next, we dive into vertical scaling techniques. After that, we present the optimizations for communications in distributed ML.
Then we continue with the communication topologies and synchronization models. Finally, we conclude this section by the discussion of the existing distributed ML frameworks.

As a side note, we use \textit{client} and \textit{participant} interchangeably in the rest of the paper.

\subsection{Introduction to Distributed ML}
Distributed ML is proposed to utilize distributed and heterogeneous computing systems to 
 solve large and complex problems where a solution cannot be obtained by a single standalone homogeneous computing device.
Distributed ML
offers two different approaches. The first is to use heterogeneous resources available 
in a single computing system such as Graphical Processing Units (GPUs). This is called vertical scaling.
The second is to use multiple machines to solve larger problems and to support fault-tolerance. This is called horizontal scaling.

GPUs have been the most common mean of vertical scaling. Given sufficient parallelism, it has been shown that GPUs significantly accelerate training \cite{cuDNN, tpus}.
For instance, NVIDIA GPUs have been popular in accelerating ML \cite{cuDNN, nvidia1}.  
Vendors such as Google have implemented their own specific hardware accelerators. Tensor Processing Units (TPUs) \cite{tpus}
are designed specifically for this purpose. Others such as Graphcore \cite{Graphcore} and SambaNova \cite{SambaNova} have followed this trend with sophisticated dataflow-based hardware designs and powerful system software tool-chains.

In contrast to vertical scaling, horizontal scaling corresponds to distributed training and inference across multiple machines. Horizontal scaling enables ML solutions to handle applications and data that do not fit in the resources of a single machine.
Additionally, the usage of multiple machines typically accelerate training and inference. 

\subsection{Parallelisms in Distributed Training and Inference}
There are three types of parallelisms used in distributed training. These are data, model and
pipeline parallelism.

\begin{figure*}[ht] 
  \centering
  \includegraphics[width=0.7\linewidth]{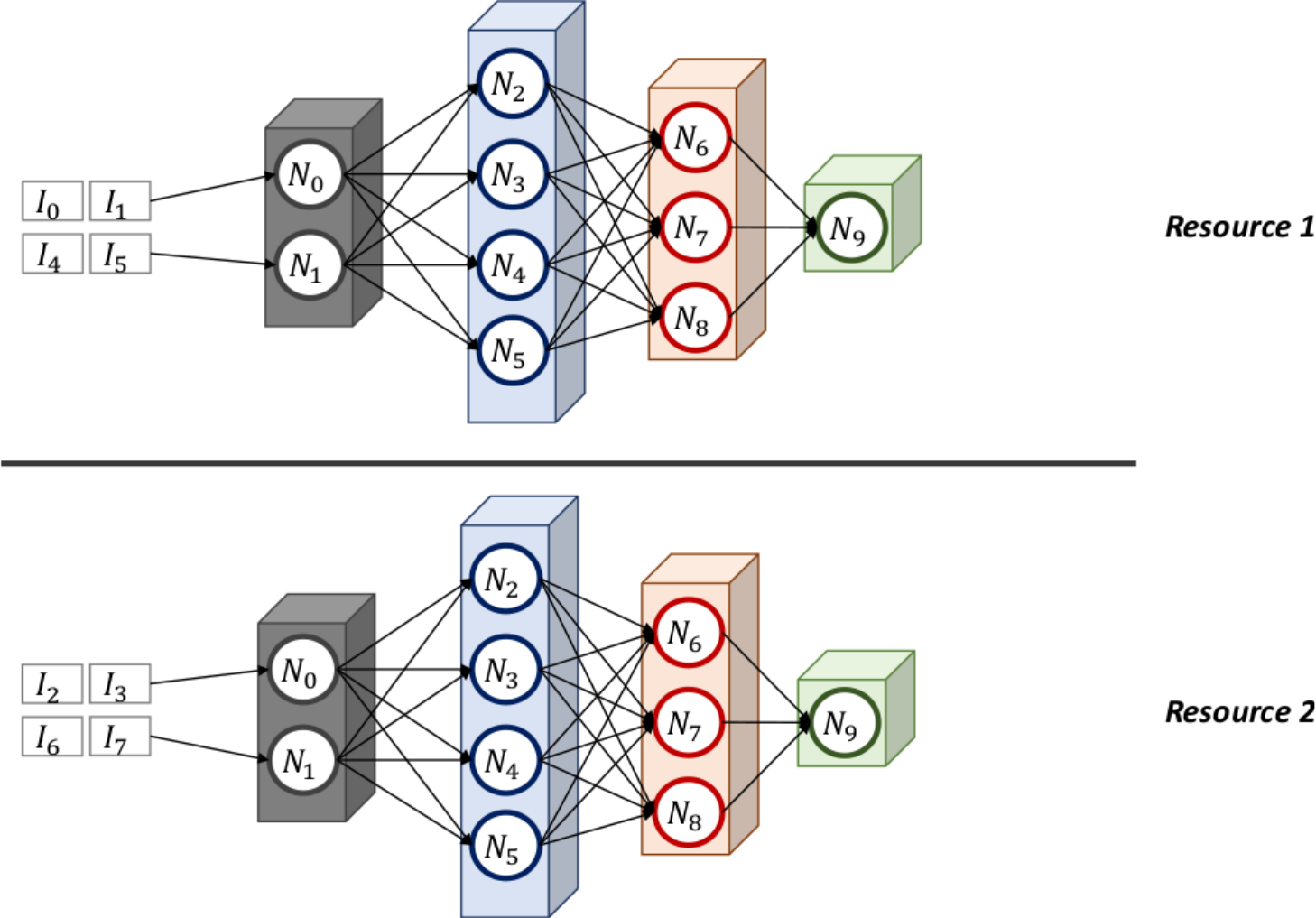}
  \caption{Data parallelism for a deep neural network.}
  \label{fig:dataParallel}
\end{figure*}

\subsubsection{Data Parallelism}
In data parallelism,
the same ML model is trained with different subsets of the data in parallel at different computing resources. Once all computing resources finish the assigned training, the models are accumulated and an average model is obtained. Then, this average model is distributed back to each computing resource for the subsequent rounds of training.
Figure \ref{fig:dataParallel} depicts data parallelism with two parallel resources.

The main advantage of data parallelism is that it is applicable to any distributed ML model without requiring expert/domain knowledge. It is also very scalable for 
 compute-intensive models, such as CNNs. 
One disadvantage of data parallelism is that model synchronization may become a bottleneck.
Another disadvantage occurs when the model does not fit in the memory of a single device.

\subsubsection{Model Parallelism}
\begin{figure*}[ht] 
  \centering
  \includegraphics[width=0.7\linewidth]{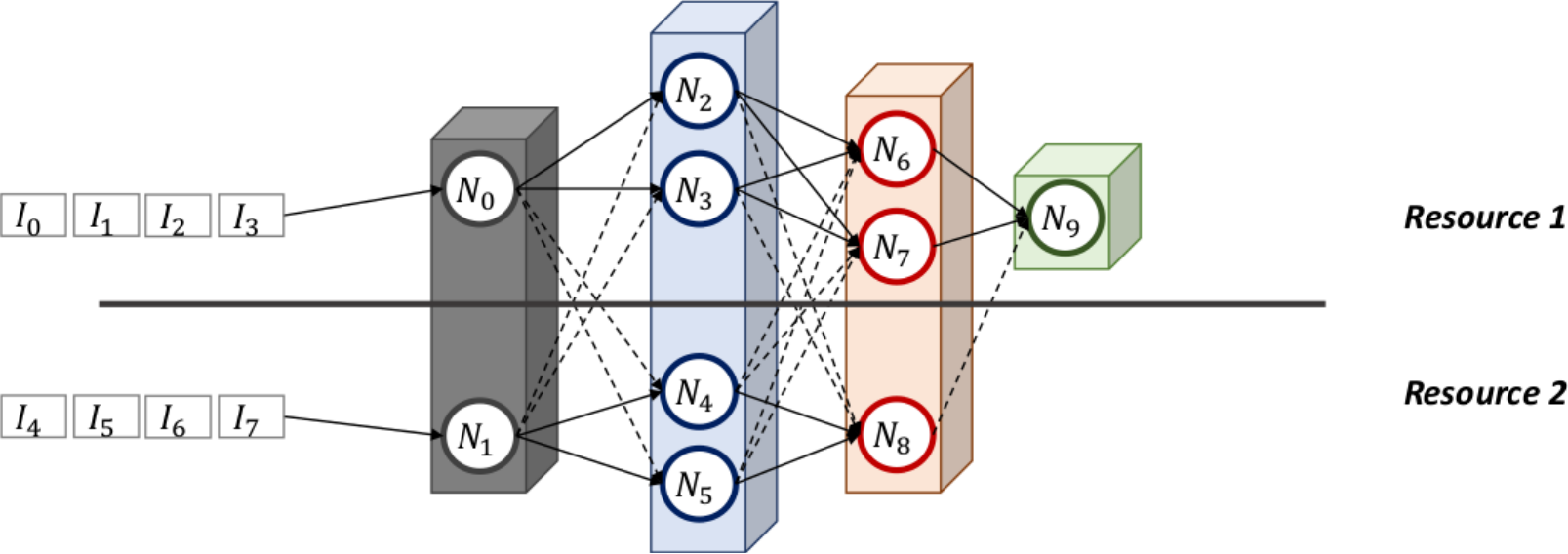}
  \caption{Model parallelism for a deep neural network.}
  \label{fig:modelParallel}
\end{figure*}

In model parallelism, the model is partitioned and distributed to different computing resources. The data is distributed as well according to the model distribution. When there is a dependency among the computing resources, synchronization is needed for the parameters (weights) to be shared consistently.  
Figure \ref{fig:modelParallel} shows model parallelism where two resources are used. An important note is that in the figure, every time that a dashed line crosses a resource boundary, at least one synchronization event must take place to ensure data consistency.

The main advantage of model parallelism is that models take less memory in each single resource (device).
Its main disadvantage is that the model partitioning is often nontrivial. Another disadvantage is the potential intensive communications among the resources. 

\subsubsection{Pipeline Parallelism}
\begin{figure*}[ht] 
  \centering
  \includegraphics[width=0.7\linewidth]{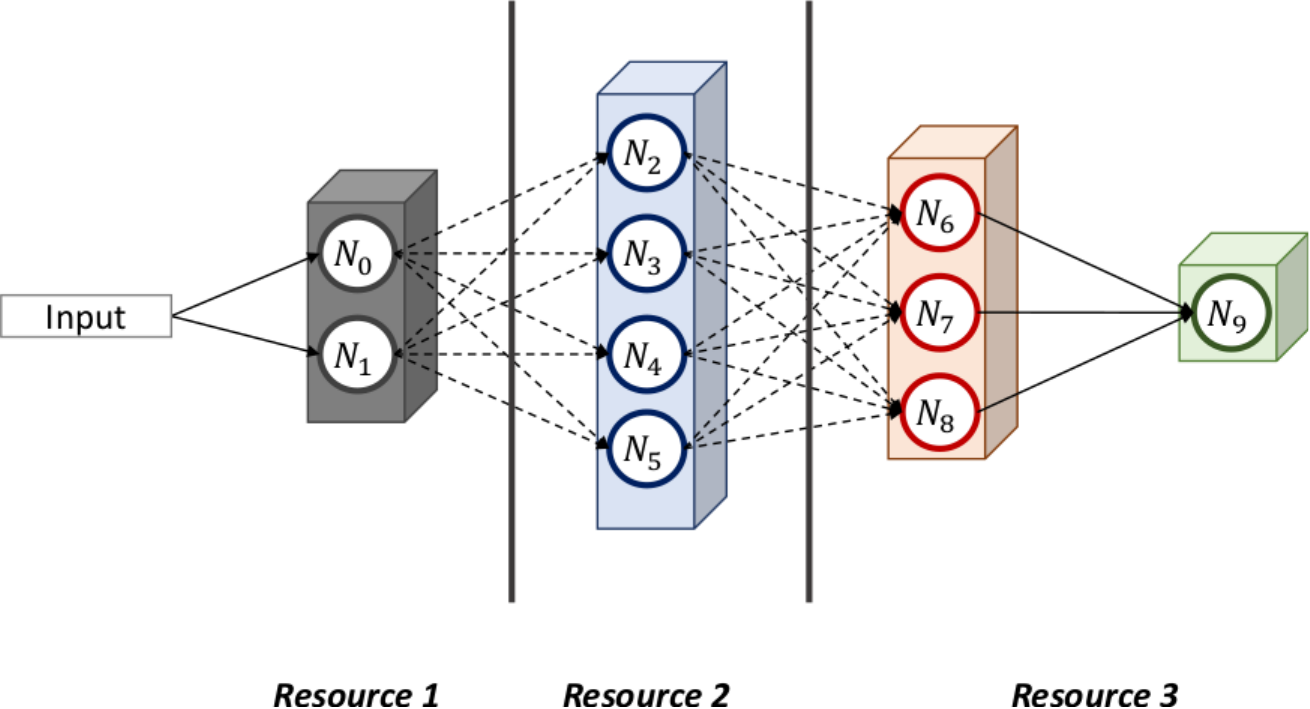}
  \caption{Pipeline parallelism for a deep neural network.}
  \label{fig:pipeParallel}
\end{figure*}

Pipeline parallelism combines model and data parallelisms.
It distributes the model and data in a such a way that there is a pipeline among the computing resources in which each resource has a different part of the model.
Pipeline parallelism maintains the advantages of model parallelism while increasing the resource utilization.
Figure \ref{fig:pipeParallel} illustrates pipeline parallelism.

\begin{figure*}[ht]
\centering 
    \subfloat[Centralized]
    {\includegraphics[width=0.3\textwidth]
    {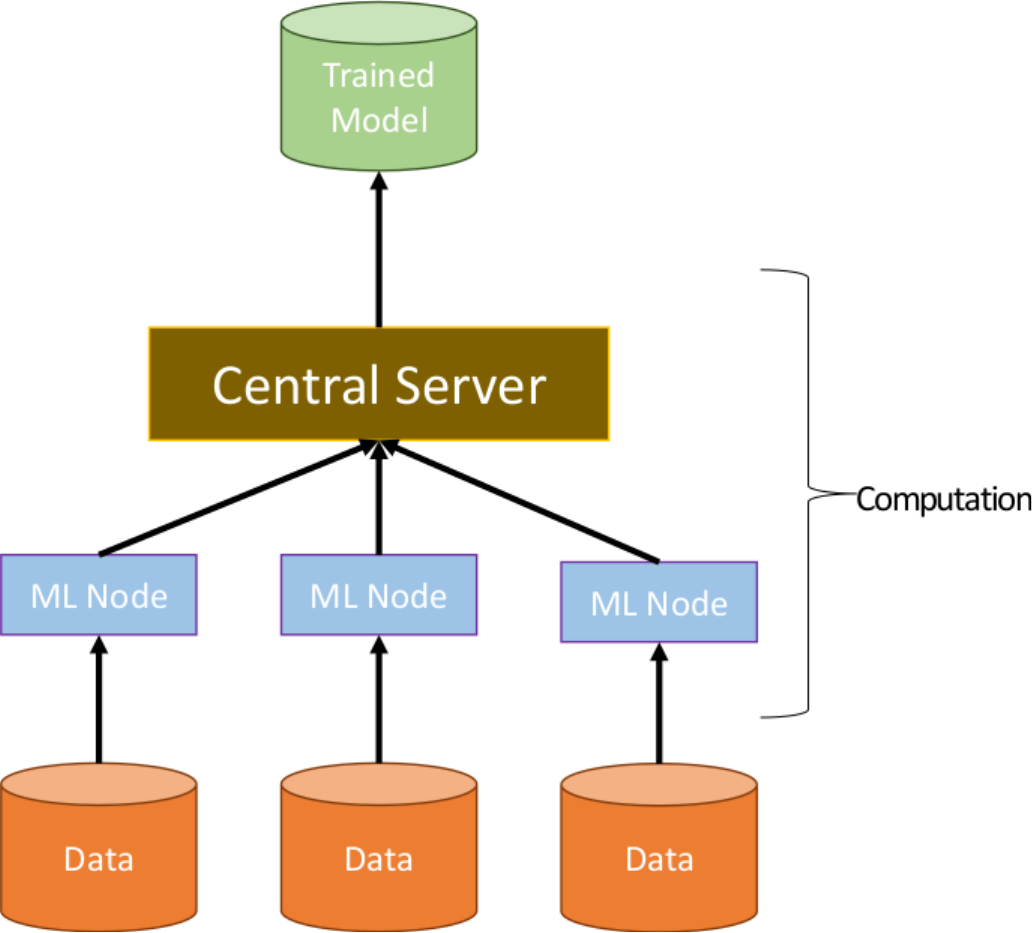}}
    \subfloat[Hierarchical]
    {\includegraphics[width=0.3\textwidth]
    {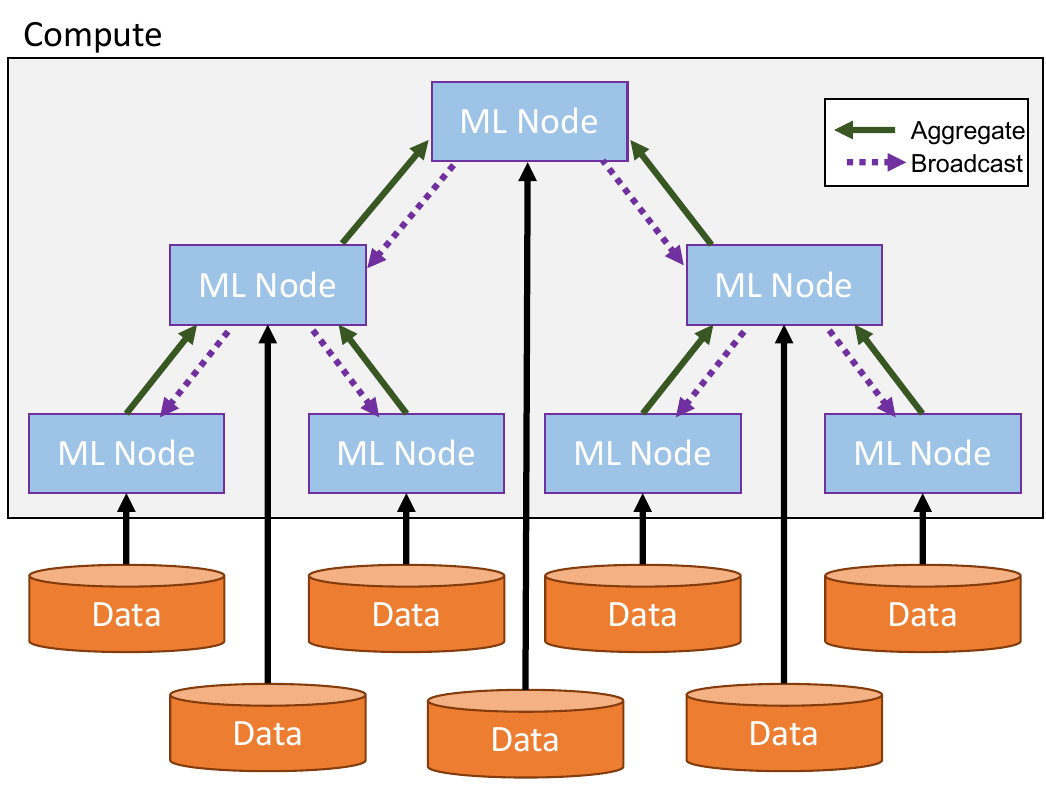}}
    \subfloat[Distributed]
    {\includegraphics[width=0.3\textwidth]
    {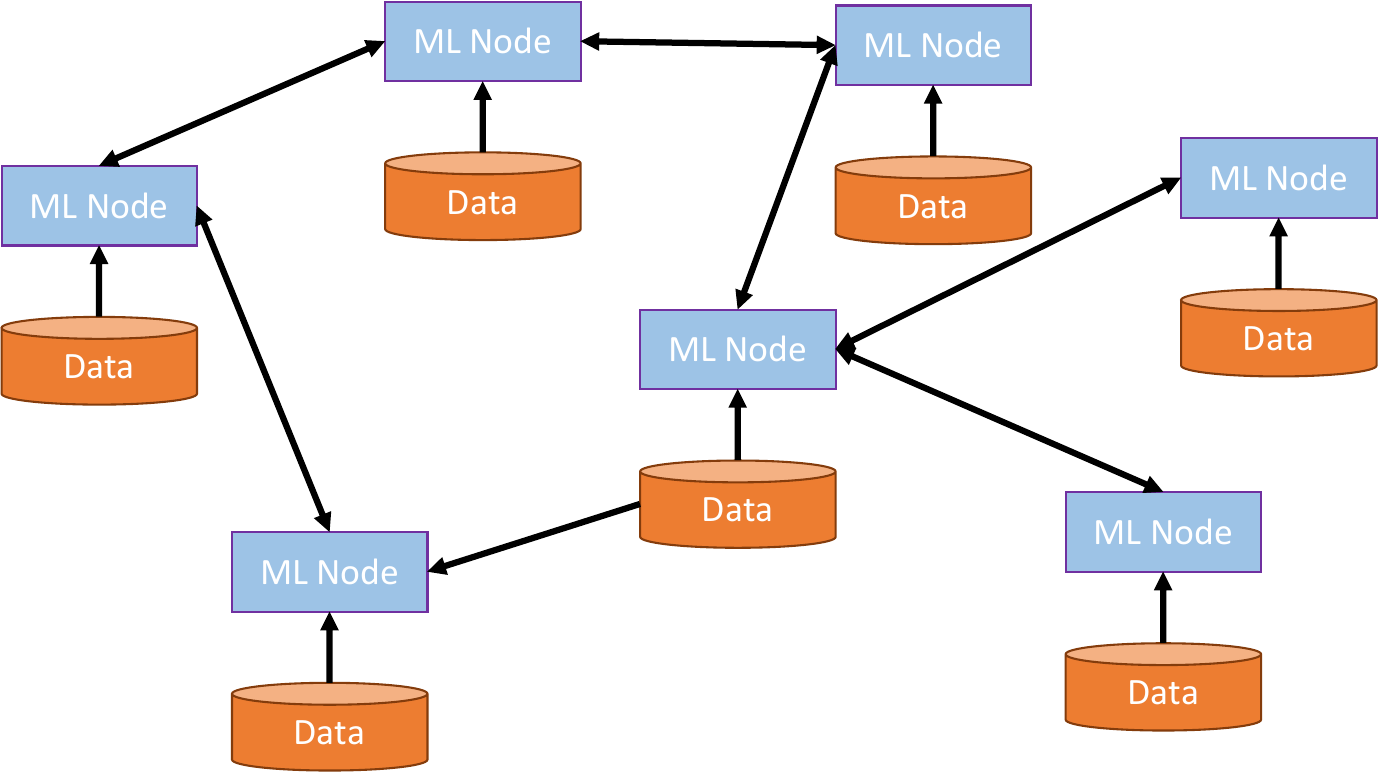}}
    \caption{Different topologies of distributed ML.}
    \label{fig:mlaggregation}
\end{figure*}

\subsection{Vertical Optimization Approaches}
We have discussed the types of parallelisms used in distributed ML above. Now, we explore three vertical optimization approaches. They are model simplification, optimization approximation, and communication optimization approaches.

\subsubsection{Model Simplification}
Model simplification refers to the reformulation of a target model to decrease its computational complexity as a way of achieving efficiency. Model simplification can be further divided into categories based on the type of the ML models.
These models can be based on kernels, trees, graphs and deep neural networks. Table \ref{tab:modelsimp}
summarizes the model simplification techniques.

\begin{table*}[ht]
    \centering
    \begin{tabular}{|l|l|l|}
    \hline
     Model Type    &   Techniques  & Existing Work  \\ \hline
    Kernel-based Models     &  
    \begin{tabular}{@{}l@{}}
    Sampling-based \\
    Projection-based    
    \end{tabular}
    &   \cite{kumar2012sampling, bouneffouf2015sampling, martinsson2011randomized, rahimi2007random}    \\ \hline
    Tree-based Models     & 
    \begin{tabular}{@{}l@{}}
    Rule sampling \\
    Feature sampling   
    \end{tabular}
    &  \cite{deng2011fast, ben2010streaming, chen2016xgboost}    \\ \hline
    Graph-based Models  &  
    \begin{tabular}{@{}l@{}}
    Sparse graph construction \\
    Anchor graph based optimization 
    \end{tabular}
    & 
    \cite{liu2010large, zhang2013fast, wang2017survey, wang2017learning, wang2016scalable} \\ \hline
    Deep Neural Network Models    &  
    \begin{tabular}{@{}l@{}}
    Efficient activation functions \\
    Filter factorization and grouping
    \end{tabular}
    & 
    \cite{nair2010rectified, maas2013rectifier, liew2016bounded, howard2017mobilenets, krizhevsky2017imagenet} \\ \hline
    \end{tabular}
    \caption{Model simplifications for different ML models.}
    \label{tab:modelsimp}
\end{table*}

Simplifications for kernel-based models are made by sampling-based or projection-based approximations. While sampling-based methods \cite{kumar2012sampling, bouneffouf2015sampling} approximate kernel matrices by random samples, projection-based methods \cite{martinsson2011randomized, rahimi2007random} use Gaussian or sparse random projections to map the data features to low dimensional sub-spaces.

Performance and scalability improvements for tree-based models, such as decision trees and random forests, are commonly based on rule \cite{deng2011fast} or feature sampling \cite{chen2016xgboost} \cite{ben2010streaming}.

Graph-based simplifications are developed for graph-based models where nodes represent the data instances and edges represent the similarity between the instances. In these models, the cost of training comes from two main sources: graph construction and the label matrix inversion.
For sparse graphs, graph construction constitutes the main cost of training. This is 
because when label propagation is used, it lowers the cost of the inversion of the label matrix and it becomes less costly than graph construction. As a result,  graph construction dominates the main computational cost.
To construct sparse graphs \cite{liu2010large}, hashing methods \cite{zhang2013fast} \cite{wang2017survey} are often used. 
Different than sparse graph models, there are also graph models that are built by anchor graphs \cite{wang2017learning}. An anchor graph is a hierarchical representation of a target graph. It is built with a small subset of the instances. This small subset is used to retain the similarities between all instances. In such a representation, the label matrix inversion is the main cost of training.
To reduce the cost of the matrix inversion, the pruning of anchors' adjacency \cite{wang2016scalable} is a common technique.

Performance improvements for deep neural networks can be achieved in two different ways. First, activation functions, such as Rectified Linear Unit (ReLU) \cite{nair2010rectified} and its variants \cite{maas2013rectifier} \cite{liew2016bounded}, can be employed instead of the expensive functions, such as sigmoid and tanh, which use the exponential function. Other techniques, specifically for CNNs, involve depth-wise filter factorization \cite{howard2017mobilenets} and group-wise convolutions \cite{krizhevsky2017imagenet}.

\subsubsection{Optimization Approximation}
Optimization approximation is a family of techniques that are 
used to reduce the cost of the optimization related
computations, i.e., gradient computations, for training. It is generally realized by computing the gradients with a small number of instances or parameters instead of all instances or parameters.
Care has to be taken since such approximations can lead to longer convergence times, local extrema, or even non-convergence. Optimization approximation can be categorized based on the specific optimization algorithm that is being used: Mini-batch gradient descent, coordinate descent, and numerical integration based on Markov chain Monte Carlo. Table \ref{tab:optim_approx} shows the existing techniques.

\begin{table*}[ht]
    \centering
    \begin{tabular}{|l|l|l|} 
    \hline
     Categories    &   Techniques  & Existing Work  \\ \hline
    Mini-batch gradient descent      &  
    \begin{tabular}{@{}l@{}}
    Adaptive sampling \\
    Adaptive learning rates \\
    Gradient corrections \\
    \end{tabular}
    &  
    \begin{tabular}{@{}l@{}}
    \cite{gopal2016adaptive}, \cite{alain2015variance}, \cite{goyal2017accurate} \\
    \cite{bengio2012practical} 
     \cite{reddi2019convergence} \\ \cite{duchi2011adaptive} 
     \cite{zeiler2012adadelta} \\
    \cite{qian1999momentum}, \cite{nesterov2013gradient} \\ \cite{sun2019survey}, \cite{le2011optimization}
    \end{tabular}
    \\ \hline
    Coordinate gradient descent  & 
    \begin{tabular}{@{}l@{}}
    Rule sampling \\
    Feature sampling 
    \end{tabular}
    & 
    \begin{tabular}{@{}l@{}}
    \cite{bayer2017generic}  \cite{yuan2012recent} \\ \cite{nutini2015coordinate} \cite{nesterov2012efficiency} 
    \cite{shi2016primer} \\ 
    \cite{yun2011coordinate}  \cite{nesterov2012efficiency}  \cite{li2015accelerated} \cite{boyd2011distributed}
    \end{tabular}
    \\ \hline
    Bayesian optimization  & 
    \begin{tabular}{@{}l@{}}
    Sparse graph construction \\
    Anchor graph based optimization
    \end{tabular}
    & 
   \cite{chib1995understanding}, \cite{griffiths2004finding}, \cite{ahn2012bayesian}  \\ \hline
    \end{tabular}
    \caption{Optimization approximation based techniques.}
    \label{tab:optim_approx}
\end{table*}

Techniques that are used for mini-batch gradient descent approximations are adaptive sampling of mini-batches, adaptive learning rates, and the improvements in gradient approximations. 
Adaptive sampling \cite{gopal2016adaptive} \cite{alain2015variance} for mini-batches takes the data distribution and gradient contributions into account rather than just using random batches of samples or making a gradual increase in the batch size \cite{goyal2017accurate}.
Learning rates are also crucial in terms of achieving fast convergence \cite{bengio2012practical}.
Adaptive learning rates can boost the speed and quality of convergence \cite{reddi2019convergence}. Further adaptive adjustments are shown to be effective  \cite{duchi2011adaptive} \cite{zeiler2012adadelta}. 
Complementary to adaptive sampling or adaptive learning rates, reducing the variance of gradients and computing more accurate gradients are shown to be effective and efficient in achieving fast convergence. Such methods use average gradients or look-ahead corrections of gradients \cite{qian1999momentum} \cite{nesterov2013gradient}. 
In addition to the accurate first-order gradients, higher-order gradients may be needed due to ill-conditioning \cite{sun2019survey} \cite{le2011optimization}. Hessian matrices are estimated by the high-order gradients to make convergence possible  \cite{sun2019survey}. 

Coordinate gradient descent are targeted at the problems where the instances are high dimensional, such as recommender systems \cite{bayer2017generic} and natural language processing \cite{yuan2012recent}. To speed up the optimizations performed by coordinate gradient descent, a small number of parameters can be selected at each iteration. Random selection of parameters has shown to be effective \cite{nutini2015coordinate} 
\cite{nesterov2012efficiency}. Parameter selection can also be based on the first and/or second-order gradients information \cite{shi2016primer} \cite{yun2011coordinate}. Another approach for speedup is to use extrapolation steps during the optimization phase \cite{nesterov2012efficiency}.  
If the optimization problem is non-convex, then studies such as \cite{li2015accelerated} \cite{boyd2011distributed} present specific solutions. For instance, Li and Lin \cite{li2015accelerated} propose 
an extended variant of accelerated proximal gradient method.

Finally, Bayesian optimization methods are commonly based on Markov chain Monte Carlo \cite{chib1995understanding} \cite{griffiths2004finding}.
Such methods employ stochastic mini-batches due to the high cost of the acceptance tests \cite{ahn2012bayesian}. 

\subsubsection{Communication Optimization Approaches}
Optimizations to reduce communication costs constitute another option to those for computation.
In these optimizations, compression of gradients is one of the two main ideas.
Some studies compress each gradient component to just 1 bit \cite{seide20141}. Others map
gradients to a discrete set of values \cite{alistarh2017qsgd} or sketch gradients into buckets and then encode them \cite{wen2017terngrad}. 
Some proposals only communicate gradients that are bigger than a certain threshold \cite{lin2017deep}.
A combination of gradient compression and low-precision learning has been shown to further reduce the communication costs \cite{zhang2017zipml}.
The other main idea for the optimization of communication is gradient delaying \cite{ho2013more}.
Ho et. al. explore the usage of gradient delays for stale synchronous parallel communications.
Zheng et. al. \cite{zheng2017asynchronous} on the other hand compute approximate second-order gradients and overlap these computations with the delays to enhance the communication efficiency. Zhang, Choromanska, and LeCun \cite{zhang2015deep} define an elastic relationship between the local and global model to avoid local minima as gradient transfers are delayed. Different than these studies, McMahan and Streeter \cite{mcmahan2014delay} introduce communication optimizations  for online learning. 

Table \ref{tab:comm_omp} summarizes these techniques.

\begin{table*}[ht]
    \centering
    \begin{tabular}{|l|l|l|}
    \hline
     Categories    &   Techniques  & Existing Work  \\ \hline
    Communications    &  
    \begin{tabular}{@{}l@{}}
  Gradient compression \\
  Gradient delay
    \end{tabular}
    &   
    \begin{tabular}{@{}l@{}}
    \cite{seide20141}, \cite{alistarh2017qsgd}, \cite{wen2017terngrad} \cite{lin2017deep}, \cite{zhang2017zipml} \\
    \cite{ho2013more}, \cite{zheng2017asynchronous}, \cite{zhang2015deep}, \cite{mcmahan2014delay} 
    \end{tabular}
    \\ \hline
    \end{tabular}
    \caption{Communication optimization approaches.}
    \label{tab:comm_omp}
\end{table*}

\subsection{Communication Topology}
In a distributed ML system, the computing resources (clusters) can be structured in
different ways. The types of topologies that the resources use can be categorized into three: centralized, hierarchical, and fully distributed (decentralized). Figure \ref{fig:mlaggregation} depicts these topologies. Table \ref{tab:topologies} summarizes our discussion.

\subsubsection{Centralized Topology} In this topology, the computation of the global model parameters, gradient averaging and communications with the distributed nodes/clients are performed at a central server. Every distributed client directly communicates with the central server and works with its local data only. A major disadvantage of a centralized topology is that the central server constitutes a single point of failure and a computational bottleneck. Advantages of a centralized topology are the ease of its implementation and inspection.
Figure \ref{fig:mlaggregation} (a) presents an example of this topology. 

\subsubsection{Hierarchical Topology} The computations and aggregation of the global model parameters are performed in a stage-wise and hierarchical way.
Each child node only communicates with its parent.
These topologies offer higher scalability than the centralized counterparts and easier manageability than the distributed counterparts.
Figure \ref{fig:mlaggregation} (b) depicts a hierarchical topology.

\subsubsection{Fully Distributed Topology} Every participant maintains a local copy of the global model in a fully distributed topology. Participants directly communicate with each other. Compared to the centralized and hierarchical topologies, scalability is much higher and the single points of failure are eliminated. However, the implementation of these topologies is relatively more complex.
Figure \ref{fig:mlaggregation} (c) shows this topology.

\begin{table*}[ht]
    \centering
    \begin{tabular}[width=\textwidth]{|l|c|c|c|c|c|} \hline
     Topology &   Complexity  & Scalability & Manageability & Single Point Failures  & Latency   \\ \hline
    Centralized    &  Low  & Low & High & Yes  & Low   \\ \hline
    Hierarchical    & Medium & Medium & Medium & Yes  & Medium   \\ \hline
    Fully Distributed  & High  & High  & Low  &  No & High  \\ \hline
    \end{tabular}
    \caption{Comparison of different communication topologies.}
    \label{tab:topologies}
\end{table*}

\subsection{Synchronization Models}
Synchronization models are techniques to guide and perform synchronization between parallel computations and communications.
These models seek to establish the best trade-off between fast updates and accurate models. To do fast updates, lower levels of synchronization are required.
In comparison, to obtain accurate models, higher levels of synchronization are needed.

As far as ML is concerned, stochastic gradient descent is one of the most popular algorithms for the optimization during the training phase. As discussed below, 
 variants of stochastic gradient descent have been implemented in accordance with the underlying synchronization model. Therefore, those variants constitute practical examples for the corresponding synchronization model.

\subsubsection{Bulk Synchronous Parallel}
It is a synchronization model \cite{valiant1990bridging} where synchronization happens between each computation and communication phase. Since this model is serializable by construction, the final output is guaranteed to be correct.
However, when there are discrepancy between the progress of parallel workers, the faster workers have to wait for the slower ones. This can 
result in significant synchronization overhead.

\subsubsection{Stale Synchronous Parallel}
This synchronization model \cite{ho2013more} allows the faster workers continue with their version of data for an additional but limited number of iterations to reduce the synchronization overheads due to the wait on the slower workers.
While this can help reduce the overheads, data consistency and  model convergence may become difficult to establish.

\subsubsection{Approximate Synchronous Parallel}
In this model, synchronization is sometimes omitted or delayed to reduce the overheads.
However, the accuracy and consistency of a model may deteriorate if care is not taken.
An advantage of approximate synchronicity is that when a parameter update is insignificant, the server can delay synchronization as much as possible.
A disadvantage is that selecting which updates are significant or not is typically difficult to do.
As an example of the application of this model, Gaia \cite{hsieh2017gaia} is an approximate synchronous parallel ML system.

\subsubsection{Asynchronous Parallel}
This synchronization model omits all synchronizations among the workers.
While these omissions may significantly reduce the computation time and communication overhead, asynchronous communications may cause ML models to produce incorrect outputs.
To give an example application,
HOGWILD algorithms \cite{desa2015taming} are developed based on asynchronous communications.

\subsection{Existing Distributed Learning Frameworks}
\label{existingframes}
There are many ML frameworks that provide distributed ML algorithms and utilities. The most popular distributed implementations are Tensorflow \cite{tensorflow2015-whitepaper, tensorflowdist, kerasdist}, PyTorch \cite{NEURIPS2019_9015, pytorchdist},  MXNet \cite{MXNet, mamidala2018mxnetmpi},  Horovod \cite{Horovod}, Baidu \cite{gibiansky2017bringing-baidu}, Dianne \cite{DIANNE}, CNTK \cite{CNTK} and Theano \cite{Bergstra10theano}.
Table \ref{tab:dist_platforms} summarizes these frameworks. 
Other than the ML frameworks above, some general-purpose distributed computing libraries, such as Apache Spark \cite{zaharia2016apache} and Hadoop \cite{hadoop}, also support distributed ML.

\begin{table*}[ht]
    \centering
    \begin{tabular}{|l|l|l|l|}
    \hline
     Frameworks    & Pros & Cons & Parallelism \\ \hline
     Tensorflow  & 
       \begin{tabular}{@{}l@{}}
  Most popular.  \\
  Strong support by Google. \\
  Efficient and scalable 
  CPU, \\
  multi-GPU, \\
  mobile implementations. \\
  Various training strategies:\\
  Multi-worker, Parameter server...
  \end{tabular} 
  & Difficult to use API
     & Data, Model \\ \hline
     PyTorch &
        \begin{tabular}{@{}l@{}}
      Dynamic computation graph \\
      Automatic differentiation \\
      Support of remote procedure calls
     \end{tabular} 
     &  
     No support for mobile
     & 
     \begin{tabular}{@{}l@{}}
     Data, Model, \\
     Pipeline 
     \end{tabular} 
     \\ \hline
      MXNet & 
      \begin{tabular}{@{}l@{}}
      High scalability \\
      Support of many languages:\\
      C++, Python, Julia, R  \\
      Usage of symbolic \\
      and imperative programming
     \end{tabular} 
      & Difficult to use API
      & Data \\ \hline
     Horovod  & 
     \begin{tabular}{@{}l@{}}
     Easy to use \\
     Supports Tensorflow, Keras, \\PyTorch, and MXNet
       \end{tabular} 
     &
     Lacks fault tolerance 
     & 
     \begin{tabular}{@{}l@{}}
     Data  Model \\
     Pipeline 
     \end{tabular} 
     \\ \hline
     Baidu & 
     Commercial ML and DL solutions 
     &
     \begin{tabular}{@{}l@{}}
     Limited scalability \\
     No support for fault-tolerance
     \end{tabular} 
     & Data, Pipeline \\ \hline
     Dianne & Java based development platform 
     & No other languages
     & Data, Model \\ \hline
    CNTK  & 
       \begin{tabular}{@{}l@{}}
    Open-source \\
    Efficient and high-performing
     \end{tabular} &
    \begin{tabular}{@{}l@{}}
    No longer actively developed \\
    Limited mobile support
     \end{tabular} & Data, Model \\ \hline
    Theano & 
   \begin{tabular}{@{}l@{}}
    Open-source and cross-platform \\
    Powerful numerical library
     \end{tabular} & Discontinued & Data \\ \hline
    \end{tabular}
    \caption{Existing distributed learning platforms.}
    \label{tab:dist_platforms}
\end{table*}

Tensorflow \cite{tensorflow2015-whitepaper} is a free and open-source software library developed for ML and DL by Google. 
In fact, Tensorflow is the most popular library among the DL libraries. It supports distributed learning with several distribution strategies, such as mirrored, multi-worker and parameter server, that are either data or model parallel \cite{tensorflowdist, kerasdist}. The library provides efficient and scalable ML implementations for CPUs, multi-GPUs and mobile devices.

PyTorch \cite{NEURIPS2019_9015} is another free and open-source  framework based on the Torch Library developed by Meta. It is a popular framework for scientific research and provides automatic differentiation and dynamic computation graphs. It supports distributed learning mainly in two ways with torch.distributed package \cite{pytorchdist}. First, same as the Tensorflow mirrored strategy, PyTorch offers distributed data-parallel training which is based on the single-program and multiple-data paradigm. Second, for the cases that do not fit into data parallelism, PyTorch provides Remote Procedure Call (RPC) based distributed training. Examples of these types of distributed training are parameter server, pipeline parallelism, and reinforcement learning with multiple agents and observers.

MXNet \cite{MXNet} is an open-source DL framework for research prototyping and production. It offers data-parallel distributed learning with parameter servers. MXNet allows mixing both symbolic and imperative programming for computational efficiency and scalability. MXNet supports many programming languages such as C++, Python, R and Julia.

Horovod \cite{Horovod} is a distributed wrapper DL framework for TensorFlow, Keras, PyTorch, and Apache MXNet. Horovod is often easy to use because it only requires an addition of a small number of library calls to the source code. Horovod supports data, model and pipeline parallelisms.

Baidu \cite{gibiansky2017bringing-baidu} was started as an easy-to-use, efficient distributed DL platform. It supports large-scale ML and can train hundreds of machines in parallel with GPUs. Baidu offers various commercial solutions, such as machine translation, recommender systems, image classification and segmentation.

Dianne \cite{DIANNE} is a distributed and ANNs-focused software framework based on OSGi which is a dynamic module system for Java. Dianne supports both model and data parallelisms and offers UI-based functionality.

The Microsoft Cognitive Toolkit (CNTK) \cite{CNTK} is open-source software for commercial-grade DL. However, it is no longer actively developed. 
It supports distributed learning through parallel Stochastic Gradient Descent (SGD)
algorithms. CNTK implements the following four parallel SGD algorithms: Data-parallel,
block momentum, model averaging, and asynchronous data-parallel SGD.

Theano \cite{Bergstra10theano} was a popular open-source Python library to define, optimize and evaluate mathematical expressions. It has support for efficient multi-dimensional arrays. Developed by Universite de Montreal, it is no longer used widely.
Theano supports data-parallel distributed learning by both synchronous and asynchronous training. It also supports multi-GPU multi-machine distributed training.

General-purpose distributed frameworks that are based on MapReduce programming model \cite{dean2008mapreduce}, such as Apache Spark \cite{zaharia2016apache} and Apache Hadoop \cite{hadoop}, supports distributed ML algorithms, applications and utilities.
Apache Spark is one of the most popular implementations of MapReduce.
It includes MLlib \cite{MLlib} which is an open-source scalable distributed ML library. MLlib consists of widely-used ML algorithms and utilities for classification, regression, clustering, and dimensionality reduction tasks.

\section{Federated Learning (FL)}
\label{federatedml}
In this section, we first introduce FL.
We then present the existing aggregation algorithms in detail.
After that, we discuss the security and privacy aspects of FL. We conclude this section by the available FL platforms and datasets.

\subsection{Introduction to FL}
FL \cite{DBLP:conf/aistats/McMahanMRHA17} is a variant of ML where training a model is done by distributed clients that individually train local models. Once local models are trained, all local model parameters are sent to a central server which then calculates the average of the parameters (weights) to compute an average model. This average model is then communicated back to the clients for subsequent local training. 
FL performs distributed training without sharing private client data. 

\begin{figure}[ht] 
  \centering
  \includegraphics[width=\linewidth]{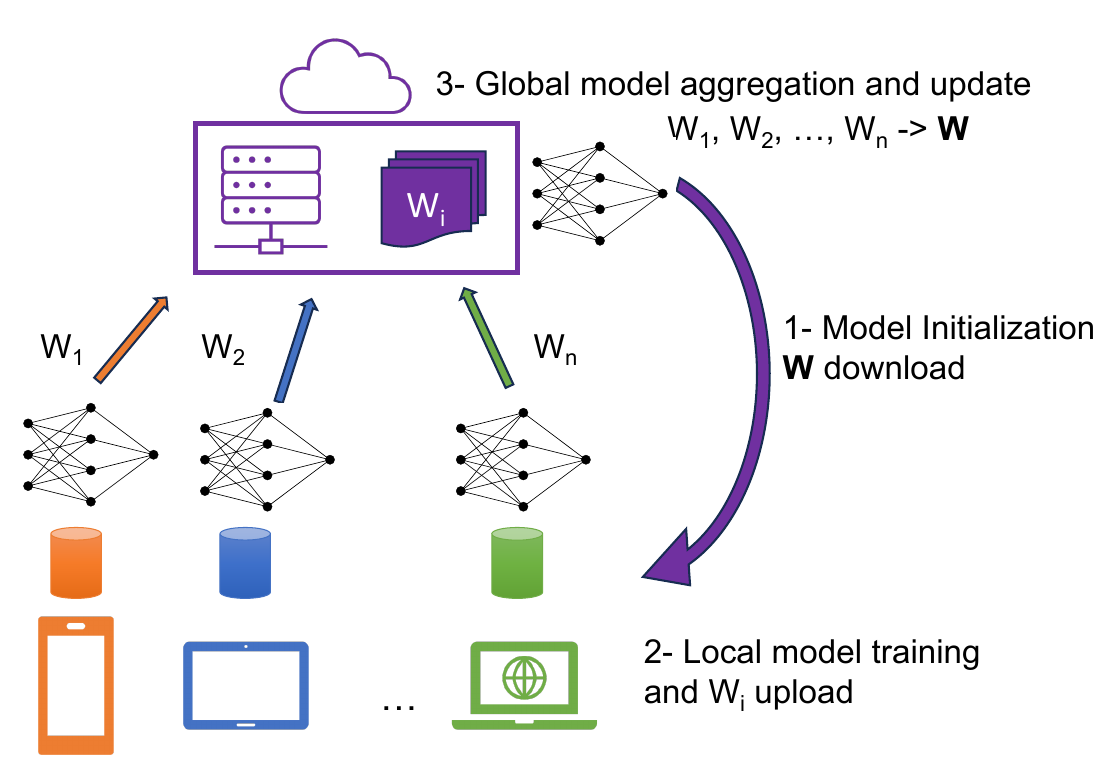}
  \caption{Federated Learning Overview.}
  \label{fig:flscheme}
\end{figure}

FL can be categorized based on how data partitioning is done.
Horizontal FL \cite{yang2019federated} refers to the case where the clients share the same feature space but have different sample spaces. This is similar to data parallelism.
An example of horizontal FL is wake-up voice recognition on smartphones. Users with different types of voices (different sample spaces) speak the same wake-up command (same feature space).

Vertical FL \cite{yang2019federated} takes place where the clients share the same sample space but have different feature spaces. As an example, the common customers (same sample space) of a bank and an e-commerce company (different feature spaces) join the training of an FL model for optimizing personal loans. 

Finally, Federated Transfer Learning \cite{liu2020secure} refers to the case where both the sample and the feature spaces are different. Federated transfer learning transfers features from different feature spaces to the same representation to train a model with the data of different clients.
An example is disease diagnosis by many different collaborating  countries with multiple hospitals which have different patients (different sample spaces) with different medication tests (different feature spaces). 

Distributed machine learning and FL have some fundamental differences. These are:
\begin{itemize}
    \item While distributed machine learning's main goal is to minimize the computational costs and achieve high scalability, FL's main goal is to provide privacy and security for the user/client data. As a result, FL is designed such that user/client data is never shared.
    \item Distributed learning assumes that the user data is independent and identically distributed (i.i.d). On the other hand, FL assumes non-i.i.d because users typically have different data distributions and types.
    \item Distributed learning is performed based on aggregating client data,  which is then distributed to different clients for training and inference. Contrarily, FL utilizes decentralized data. The client data is never shared and is never aggregated on a central server.
\end{itemize}

\begin{algorithm}[ht]
\caption{Federated learning: client and server functions}\label{alg:cap}
\begin{algorithmic}[1]
\State $i \gets isClient$ or $isServer$
\State $E \gets totalEpochs$
\State $B \gets totalNumBatches$
\State $\eta \gets learningRate$
\State $w \gets initialWeights$
\If{$i = isClient$} 
\Function{UpdateClientWeight}{w,k}~\label{lst:line:Client}
\For{epochs $e$ from 1 to $E$}
    \For{batchs $b$ from 1 to $B$}
        \State $w \gets w - \eta \nabla l(w,b)$
    \EndFor
\EndFor
\State \Return w to the server
\EndFunction
\Else
\Function{ServerUpdateWeight}{}~\label{lst:line:Server}
\State $t \gets currentRoundID$
\For{$k$ in sub batch of $K$ clients}
    \State the following is done in parallel
    \State $w^{k}_{t+1} \gets UpdateClientWeight(k,w^{k}_{t})$
\EndFor
\State $w_{t+1} \gets \frac{\sum_{k=1}^{K} w^{k}_{t+1}}{K}$
\EndFunction
\EndIf 
\end{algorithmic}
\end{algorithm}

We now discuss the very first FL algorithm, FedAvg, proposed by
McMahan et. al. \cite{DBLP:conf/aistats/McMahanMRHA17}.
Algorithm~\ref{alg:cap} describes FedAvg.
It shows the action taken by the server and clients during a round of FL. The clients train the model with their data. Once trained, the weights are sent to the server as described by the $UpdateClientWeight$ function on line~\ref{lst:line:Client}. Once the server receives the weights from the clients, which is done in parallel, it averages out all the weights and sends the average weights back to each clients, as seen in the $ServerUpdateWeight$ function on line~\ref{lst:line:Server}. Training is repeated if the data changes. This is to keep the weights updated. 

\subsection{FL Applications}
FL has a wide range of applications across different domains and settings. Some of them are:
\begin{itemize}
    \item Smartphones: FL has been used to develop ML applications for smartphones such as next-word prediction, and face and voice recognition.
    \item Healthcare: FL has been applied successfully for research problems in medical studies such as drug discovery and brain tumor segmentation.
    \item The internet of things (IoT): IoT is a network of digital or mechanical computing objects that have sensors, software, and other computing technologies. IoT exchanges data with other devices and systems over the internet to perform specific learning tasks. Applications of FL in IoT include autonomous driving and intrusion and anomaly detection.
    \item Finance: FL has been adopted to detect/identify financial crimes such as fraudulent loans and money laundering.
\end{itemize}

\subsection{FL Aggregation Algorithms}
In FL, due to data parallelism and horizontal FL, aggregation algorithms are needed to aggregate the models or gradients between the participants.
As stated above,
the very first aggregation algorithm, called Federated Averaging (FedAvg), was introduced by McMahan et. al. \cite{DBLP:conf/aistats/McMahanMRHA17} who essentially kick-started FL itself. FedAvg computes the global model parameters by averaging the parameter updates of the participants. Once the global parameters are computed and updated, these parameters are communicated back to the participants. FedAvg is a straightforward algorithm however, it is biased toward the participants who have favorable network conditions. 

Aggregation algorithms have been studied extensively for centralized topologies \cite{FedBCD, asad2020fedopt}. 
To decrease the communication overheads, Liu et. al. propose the Federated Stochastic Block Coordinate Descent (FedBCD) algorithm \cite{FedBCD} in which each participant makes multiple local updates before synchronizing with other participants. 
Differently, FedOpt \cite{asad2020fedopt} uses gradient compression to reduce communication overhead while sacrificing accuracy. 
Furthermore, for edge devices where computational resources are limited, algorithms such as FedGKT \cite{he2020group} are developed.

A significant objective in FL is to provide fairness.
Fairness means that the clients equally contribute to the global model with respect to certain metrics. Researchers have proposed algorithms such as Stochastic Agnostic Federated Learning (SAFL) \cite{AgnosticFederated} and FedMGDA+ \cite{FederatedMultiobjective} to achieve fairness.

Adaptive FL and its impact on convergence and accuracy have been explored in various recent works.
ADAGRAD \cite{reddi2020adaptive} offers an adaptive approach to ML  optimization compared to FedAvg. ADAGRAD and its variants dynamically choose server and client learning rates and momentum parameters during training. Mime Lite \cite{karimireddy2021breaking} is a closely related study where adaptive learning rates and momenta are reported to improve accuracy.

Some recent aggregation algorithms support heterogeneity of participant data. FedProx \cite{FedProx} is such an algorithm used for FL over heterogeneous data and resources. SCAFFOLD \cite{SCAFFOLD} is another algorithm that accounts for heterogeneous data while reducing the number of rounds to converge. 
FedAtt \cite{Attentive} accounts for the client contributions by attending to the importance of their model updates. The attention is quantified by the similarity between the server model and the client model in a layer-wise manner.
FedNova \cite{wang2020tackling}  proposes a normalized averaging method as a way to avoid objective inconsistencies and to achieve fast convergence for highly heterogeneous clients.

Personalization is another important consideration in FL.
There has been extensive research on personalized FL \cite{deng2020adaptive} \cite{shamsian2021personalized}. 
Tan et. al. \cite{tan2022towards} offer a survey of the latest  personalization techniques. 

When the topology of the clients in FL is hierarchical, an aggregation algorithm needs to take the hierarchy into account. Numerous hierarchical aggregation algorithms have been proposed for such settings \cite{ yurochkin2019statistical, yurochkin2019bayesian}.  
Among hierarchical solutions, SPAHM \cite{yurochkin2019statistical} and PFNM \cite{yurochkin2019bayesian} are Bayesian FL methods. 
Similarly, for decentralized topologies, decentralized algorithms have been developed \cite{ hegedHus2021decentralized, ye2022decentralized}.  

Considering fault-tolerance in FL, Krum \cite{blanchard2017machine} is an aggregation scheme that is reportedly resilient to Byzantine failures \cite{lamport2019byzantine} where computing processes fail arbitrarily and failure symptoms are different for different observers. For these types of failures, more fault-tolerant FL studies are needed.

\subsection{Security and Privacy in FL}
\begin{table*}
    \centering
    \begin{tabular}{|l|l|l|}
    \hline
   Defense Type & Addressed Attacks & Potential Negative Effects  \\ \hline
   \begin{tabular}{@{}l@{}}
   Differential privacy \\
   \cite{el2022differential, dwork2006differential, dwork2014algorithmic, wei2020federated, truex2020ldp} 
   \end{tabular} 
   & 
   \begin{tabular}{@{}l@{}}
  Data  Poisoning \cite{tolpegin2020data, 10.1145/3510548.3519372, 9618642} \\
  Model  Poisoning \cite{panda2022sparsefed, cao2022mpaf, 9762272} \\
  Inference attacks \cite{MOTHUKURI2021619, nasr2019comprehensive, liu2022threats} 
    \end{tabular} 
    & Decreased model utility \\ \hline
     \begin{tabular}{@{}l@{}}
     Homomorphic encryption \\
     \cite{zhang2020batchcrypt, hardy2017private} 
     \end{tabular}
     & 
   \begin{tabular}{@{}l@{}}
  Inference attacks \cite{MOTHUKURI2021619, nasr2019comprehensive, liu2022threats} \\
  GAN-based attacks \cite{manoharan2022svm, zhang2020poisongan, rosenberg2021adversarial}
    \end{tabular} 
    &
     High computational costs \\ \hline
 \begin{tabular}{@{}l@{}}
 Trusted execution environments \\
 \cite{mo2021ppfl, chen2020training} 
 \end{tabular}
 & 
   \begin{tabular}{@{}l@{}}
  Inference attacks \cite{MOTHUKURI2021619, nasr2019comprehensive, liu2022threats} \\
  Model Poisoning \cite{panda2022sparsefed, cao2022mpaf, 9762272}  \\
    \end{tabular} 
    &
     Specialized hardware \\ \hline
      \begin{tabular}{@{}l@{}}
 Secure Multi-party Computation \\
 \cite{mugunthan2019smpai}
 
  \end{tabular} & 
   \begin{tabular}{@{}l@{}}
  GAN-based attacks \cite{manoharan2022svm, zhang2020poisongan, rosenberg2021adversarial} \\
  Inference attacks \cite{MOTHUKURI2021619, nasr2019comprehensive, liu2022threats} \\
Eavesdropping \cite{MOTHUKURI2021619, liu2022threats}
    \end{tabular} 
    &
High computational costs \\ \hline

     Blockchain  \cite{qu2022blockchain, lu2019blockchain} &   
    Blockchain attacks \cite{saad2020exploring, chen2022survey} &
    High resource costs \\ \hline

     Data anonymization \cite{choudhury2020anonymizing} &   
     \begin{tabular}{@{}l@{}}
  GAN-based attacks \cite{manoharan2022svm, zhang2020poisongan, rosenberg2021adversarial} \\
  Inference attacks \cite{MOTHUKURI2021619, nasr2019comprehensive, liu2022threats} \\
    \end{tabular} 
    &
    Decreased data usability \\ \hline

     Anomaly detection \cite{li2019abnormal} &   
     \begin{tabular}{@{}l@{}}
 Data Poisoning \cite{tolpegin2020data, 10.1145/3510548.3519372, 9618642} \\
  Model Poisoning \cite{panda2022sparsefed, cao2022mpaf, 9762272} \\
  Free-riders \cite{MOTHUKURI2021619, liu2022threats}
    \end{tabular} 
    &
    Detection latency  \\ \hline
    
    \end{tabular}
    \caption{Attacks and defenses in FL.}
    \label{tab:attacksdefenses}
\end{table*}

The security of FL entails ensuring the triad of confidentiality, integrity and availability of its data and models, and particularly, data privacy. Privacy is defined as the protection of the raw data against the information leakage. In this section, we first summarize the attack types and then the defensive actions and methods existing in the FL literature \cite{MOTHUKURI2021619}.

\subsubsection{Attacks} 

There are numerous attack types in FL.
Poisoning attacks aim to tamper with and/or alter the data or the model. Data poisoning \cite{tolpegin2020data, 10.1145/3510548.3519372, 9618642} refers to altering the features in the training data or generating false data to degrade the performance of a model on the unseen data. Model poisoning \cite{panda2022sparsefed, cao2022mpaf, 9762272} refers to the modification of the model parameters and/or the fabrication of false weights that are communicated between the participants and the servers.

 Backdoor attacks \cite{bagdasaryan2020backdoor, gong2022backdoor} inject malicious instructions into the models while not impacting their expected performance. These attacks are non-transparent and notoriously difficult to detect.

Inference attacks \cite{MOTHUKURI2021619, nasr2019comprehensive, liu2022threats} involve gaining knowledge of the sensitive information of the participants, the training data or the model through the communications occurring during training or inference. Membership inference attacks aim to learn if a sample has been used as a training instance. Property inference attacks aim to learn the meta-characteristics of the training data. Class representative inference attacks aim to learn representative samples of a target class.

Generative Adversarial Networks (GANs) based attacks \cite{manoharan2022svm, zhang2020poisongan, rosenberg2021adversarial} are used to launch poisoning attacks where GANs generate the altered or false data and/or model parameters.

 There are many other attack types \cite{rigaki2020survey, goldblum2022dataset}, such free-riders \cite{MOTHUKURI2021619, liu2022threats} and Eavesdropping \cite{MOTHUKURI2021619, liu2022threats}.

\subsubsection{Defenses} The most commonly used attack defense mechanisms can be categorized by the usage of trusted execution environments \cite{mo2021ppfl, chen2020training}, homomorphic encryption \cite{zhang2020batchcrypt, hardy2017private}, differential privacy \cite{el2022differential, wei2020federated, truex2020ldp}, and possibly some combinations of them. 
There are many other techniques which are based on GANs \cite{PDGAN}, anomaly detection \cite{li2019abnormal}, secure multi-party computation \cite{mugunthan2019smpai}, data anonymization \cite{choudhury2020anonymizing}, and blockchains \cite{qu2022blockchain, lu2019blockchain}.

A trusted execution environment \cite{mo2021ppfl, chen2020training} is an (hardware/software) architecture where the program execution is secured and information leakage is not possible. Such architectures use specialized designs to prevent unauthorized accesses as well as privacy violations in FL \cite{mo2021ppfl, chen2020training}.

 Homomorphic encryption \cite{fontaine2007survey} is a certain type of encryption in which the decryption of the results of the computations performed on the encrypted data is the same as the result of the same computations performed on the unencrypted data. Homomorphic encryption has various levels depending on the whether addition and/or multiplication is supported. It has been adapted for data privacy in FL \cite{zhang2020batchcrypt, hardy2017private}.
 
 Differential privacy \cite{dwork2006differential, dwork2014algorithmic} is a technique for achieving data privacy by adding noise to raw data. It is commonly used in FL \cite{el2022differential, wei2020federated, truex2020ldp}. 

Table \ref{tab:attacksdefenses} reviews attacks and defenses in FL. In addition, Table \ref{tab:comparisonofdefense} compares the defense mechanisms in terms of the strength of the protection, the computational and communication efficiency, robustness, scalability, and generalizability of a mechanism.

\begin{table*}
    \centering
    \begin{tabular}{|l|l|l|l|l|l|}
    \hline
    Defense Type & Protection & Efficiency & Robustness & Scalability & Generalizability \\  \hline
    \begin{tabular}{@{}l@{}}
    Differential privacy \\
    \cite{el2022differential, dwork2006differential, dwork2014algorithmic, wei2020federated, truex2020ldp}
    \end{tabular}
    & High  & High  & High  & High & High  \\  \hline
    \begin{tabular}{@{}l@{}}
    Homomorphic encryption \\
    \cite{fontaine2007survey} 
    \end{tabular}
    & High & Low & High & Low & High  \\  \hline
    
    \begin{tabular}{@{}l@{}}
    Trusted execution environments \\
    \cite{mo2021ppfl, chen2020training} 
    \end{tabular}
    & Medium &  High & Medium & Low  & High  \\  \hline
    \begin{tabular}{@{}l@{}}
    Secure multi-party computations \\ \cite{mugunthan2019smpai} 
    \end{tabular}
    & High & Medium  & High & Low & Medium \\  \hline
    
    Blockchain \cite{qu2022blockchain, lu2019blockchain} & High & Low & Medium & High & Medium \\  \hline
    Data anonymization \cite{choudhury2020anonymizing} & Medium  &  Medium & Low & High & High \\  \hline
    Anomaly detection \cite{li2019abnormal} & Medium & High & Medium  & High  & Low  \\  \hline
    \end{tabular}
    \caption{Comparison of the defense mechanisms in terms of the strength of the protection, the computational and communication efficiency, robustness, scalability, and generalizability of a mechanism.}
    \label{tab:comparisonofdefense}
\end{table*}

\subsection{Existing FL Frameworks}
The most widely used FL frameworks are
TensorFlow Federated \cite{tensorflowfed, kerasfed}, IBM Federated Learning \cite{IBMFederated}, NVIDIA FLARE \cite{NVIDIAFLARE}, FedML \cite{chaoyanghe2020fedml}, Federated AI Technology Enabler (FATE) \cite{fate}, PySyft \cite{PySyft}, and Open Federated Learning (OpenFL) \cite{openfl_citation}. Table \ref{tab:federated_platforms} summarizes the existing FL frameworks.

\begin{table*}[ht]
    \centering
    \begin{tabular}{|l|l|l|l|}
    \hline
     Frameworks & Aggregation Algorithm & Parallelism & Privacy and Security \\ \hline
     \begin{tabular}{@{}l@{}}
     TensorFlow Federated \\
     Keras Federated
     \end{tabular} & 
     \begin{tabular}{@{}l@{}}
     FedAvg, FedProx, \\
     FedSGD, Mime Lite 
     \end{tabular}
     & Data, Model & Differential privacy \\ \hline
     IBM Federated  & 
     \begin{tabular}{@{}l@{}}
     FedAvg,  SPAHM, \\
     PFNM, Krum
      \end{tabular}
     & Data, Model &  
     \begin{tabular}{@{}l@{}}
     Differential privacy \\
     Secure multi-party computation \\
     Homomorphic encryptions 
     \end{tabular} \\ \hline
      NVIDIA FLARE  & 
      \begin{tabular}{@{}l@{}}
      FedAvg, FedProx, \\
      SCAFFOLD 
       \end{tabular}
      & Data & 
      \begin{tabular}{@{}l@{}}
      Differential privacy \\
      Homomorphic encryption
      \end{tabular} \\ \hline
     FedML  & 
     \begin{tabular}{@{}l@{}}
     FedAvg, FedOpt, \\
     FedNova, FedGKT 
      \end{tabular}
     & Data, Model & 
     \begin{tabular}{@{}l@{}}
     Differential privacy \\
     Cryptography \\ 
     Coding approaches
     \end{tabular} \\ \hline
     FATE  & FedAvg & Data, Pipeline & 
      \begin{tabular}{@{}l@{}}
       Homomorphic encryption \\
       RSA
      \end{tabular} \\ \hline
    PySyft  & 
    \begin{tabular}{@{}l@{}}
    FedAvg, FedProx, \\
    FedSGD 
     \end{tabular}
    & Data, Model & 
     \begin{tabular}{@{}l@{}}
      Differential privacy \\
      Homomorphic encryption
      \end{tabular} 
    \\ \hline
    OpenFL  & 
    \begin{tabular}{@{}l@{}}
    FedAvg, \\
    FedADAGRAD 
    \end{tabular} 
    & Data & 
    \begin{tabular}{@{}l@{}}
    Trusted execution environments \\
    RSA \\
    Differential privacy
    \end{tabular} \\ \hline
    \end{tabular}
    \caption{Existing FL platforms.}
    \label{tab:federated_platforms}
\end{table*}

TensorFlow Federated \cite{tensorflowfed} (and Keras Federated \cite{ kerasfed}) is an open-source framework for FL by Google.  It enables researchers to simulate FL algorithms. FedAvg, FedProx, FedSGD, and Mime Lite are some of the FL aggregation algorithms that are readily available. TensorFlow Federated supports data and model parallelisms.
It provides differential privacy as a privacy measure. TensorFlow Federated has two main APIs. FL API offers built-in algorithms. FL Core API offers a set of lower-level functionalities for new algorithms to be implemented.

IBM Federated Learning \cite{IBMFederated} provides support for FL and DL models written in Keras, PyTorch and TensorFlow. 
FedAvg, SPAHM, PFNM, and Krum are among the available aggregation algorithms. IBM FL supports data and model parallelism. In addition, 
differential privacy, secure multi-party computation and homomorphic encryption are available defenses for ensuring privacy and security.
IBM FL also offers the implementations of several topologies and communication protocols.

NVIDIA FLARE (Federated Learning Application Runtime Environment) \cite{NVIDIAFLARE} is a modular open-source software development kit (SDK) for FL which offers secure and privacy-preserving distributed learning. FLARE provides FL algorithms such as
FedAvg, FedProx and SCAFFOLD. It offers differential privacy and homomorphic encryption. FLARE SDK has several components, such as a simulator for prototyping, secure management tools for provisioning and deployment and an API for extensions.

FedML \cite{chaoyanghe2020fedml} framework offers a wide-range of cross-platform FL capabilities including natural language processing, computer vision, and GNNs. FedAvg, FedOpt, FedNova and FedGKT are the supported FL algorithms. FedML offers defense mechanisms such as differential privacy, cryptography routines, and several coding
methods. It supports data and model parallel distributed learning.
FedML models can be trained and deployed at the edge or on the cloud.

FATE \cite{fate} is an open-source platform initiated by WeBank, a bank based in Shenzhen, China. It provides a diverse set of FL algorithms, such as tree-based algorithms, DL, and transfer learning. It offers a set of modules consisting of an ML algorithms library, a high-performance serving system, an end-to-end pipeline system, a multi-party communication network system, and a module for cloud technologies. FATE provides homomorphic encryption and RSA for secure and privacy preserving training. FATE supports data and pipeline parallelisms.  

PySyft \cite{PySyft} is an open-source multi-language library that provides secure and private DL and FL in Python for frameworks such as PyTorch, Tensorflow and Keras. It supports differential privacy and homomorphic encryption. FedAvg, FedProx and FedSGD are among the available aggregation algorithms. Training can be data or model parallel.

OpenFL \cite{openfl_citation} is an open-source Python framework originally developed by Intel Labs. It provides a set of workflows for the researchers to experiment with FL. FedAvg and ADAGRAD algorithms are built-in.  OpenFL's capabilities include trusted execution environments, RSA, differential privacy.

\subsection{FL Datasets}
As FL research progresses, new datasets are being built. One of the most well-known datasets for FL is
the LEAF \cite{LEAF}. It is a suite of open-source federated datasets. There are a total of six different datasets.
One of the datasets, called FEMNIST, is built for image classification. Sentiment140, which consists of Tweets, is a dataset for sentiment analysis. Shakespeare is a text dataset of Shakespeare Dialogues which is used for next character prediction. Celeba is an image classification dataset of celebrity images. There is a synthetic classification dataset which is generated for the FL models that are device-dependant. 
Lastly, the Reddit comments dataset is used for next word prediction.

TensorFlow Federated \cite{tensorflowfeddatasets} offers several datasets to support FL simulations. While some of its datasets are the same as those of LEAF, there are also different datasets, such as the federated CIFAR-100 dataset, the FLAIR dataset, and the federated Google Landmark v2 dataset.

Street Dataset \cite{ImageFed} is a real-world image dataset. It contains images generated from street cameras. A total of seven object categories annotated with bounding boxes. This dataset is built for object detection tasks.

CC-19 \cite{covid19fed} is a new dataset related to the latest family of coronavirus (COVID-19). It contains the Computed Tomography (CT) scan of subjects and is built for image classification.

FedTADBench \cite{FedTADBench} offers three different datasets to evaluate time series anomaly detection algorithms.

\section{Open Questions and Challenges}
\label{questionschallenges}
In this section, we summarize the challenges that ML and FL face. We only present major problems. This is because there is a large number of open problems, and we choose to keep our presentation concise and focused.
\subsection{Challenges for Parallel and Distributed ML}
The major challenges with parallel and distributed ML are related to performance, fault-tolerance, security and privacy \cite{SurveyonDistributed, pitropakis2019taxonomy, xue2020machine}.

Typically, in distributed and parallel training, additional resources are used to decrease wall-clock time \cite{alqahtani2019performance}. Such additional resources can be multiple machines, multiple GPUs and high-end communication networks. As a result, the decrease in wall-clock time may not 
 compensate for the additional resources or their energy consumption. Therefore, research studies, such as \cite{DemystifyingParallelandDistributed}, are needed to investigate this trade-off with different applications and system architectures.

Distributed and parallel ML platforms, especially those executed on high-performance computing systems, often consider fault-tolerance as a second-class concern.
However, given the sizes of the latest large-scale computing systems, failures are common; not rare \cite{qiao2019fault}. As a result, efficient checkpointing and/or replication solutions \cite{qiao2019fault} are needed to recover from errors and to limit the amount of lost computation due to a failure. 

Ensuring security and privacy for distributed and parallel ML has consistently been a serious concern \cite{pitropakis2019taxonomy}. While FL was devised for the privacy of user data, there have been many novel types of attacks \cite{rigaki2020survey, goldblum2022dataset}. These attacks include adversarial \cite{rosenberg2021adversarial}, poisoning, evasion, backdoor, and integrity attacks \cite{goldblum2022dataset, xue2020machine}. As such attacks get sophisticated, so must their defenses.
Moreover, the systematic deployment of the defenses to the physical systems as well as the evaluation of these deployments have not studied well \cite{xue2020machine}.
Furthermore, there is a lack of the rigorous efficiency and efficacy studies of attack defense mechanisms. As a result of these issues, security and privacy for ML remain an open problem.

\subsection{Challenges for FL}
The main challenges in FL are two-fold \cite{AdvancesOpenFederated}: explainability and interpretability, and federated GNNs. 
Explainability and interpretability refer to the understanding of the contributions of the clients or the data features. For instance, 
Shapley values are proposed \cite{wang2019interpret} to quantify the impact of the features on the model output.
 Zheng et. al. propose a quantified ranking of features \cite{zheng2020vertical}.
Similarly, there are studies \cite{chen2021fed} targeting vertical FL.
Several works introduce tailored measures of interpretability such as 
 \cite{li2023balancing} defining a measure based on the gradients. However, in general, the problem of explainability and interpretability remains open because 
i) ensuring privacy while building explainable models is not trivial,
ii) the aggregation of the local parameters obscures interpretability,
iii) there is a lack of datasets that are not composed of images or text, and
iv) there is a lack of a general framework for explainable federated models. 

Research for FL with GNNs 
\cite{he2021fedgraphnn, he2021spreadgnn, wang2022graphfl} has recently started.
For instance, FedGraphNN \cite{he2021fedgraphnn} provides an FL benchmark system to evaluate various graph models, algorithms and datasets. Another example is GraphFL \cite{wang2022graphfl} which is designed to classify nodes on graphs. However, many questions are still waiting to be solved, such as the 
protection against malicious attacks, interpretability, lack of modern graph neural frameworks for FL \cite{liu2022federated}.

\section{Conclusions}
\label{conclusion}
In this work, we provided a review of modern large-scale, parallel and distributed ML: the state-of-the-art algorithms, optimization methods, types of parallelisms, communication topologies, synchronization models, and the existing frameworks. Moreover, we reviewed FL.
We discussed various aggregation algorithms in FL. In addition, we reviewed the security and privacy aspects including various types of attacks and defense mechanisms. Moreover, we explored the existing FL frameworks and datasets. We concluded our study with the open research problems and challenges in large-scale distributed ML and FL. The major challenges are typically related to performance, security, privacy, explainability, portability, and fault-tolerance. 

\section*{Acknowledgment}
This work was supported by the U.S. DOE Office of Science, Office of Advanced Scientific Computing Research, under award 66150: "CENATE - Center for Advanced Architecture Evaluation" project. The Pacific Northwest National Laboratory is operated by Battelle for the U.S. Department of Energy under contract DE-AC05-76RL01830.

\bibliographystyle{unsrt}
\bibliography{references}


\end{document}